%% file: themis.tex
\documentclass[sigconf]{acmart}

\usepackage[utf8]{inputenc} 
\usepackage[T1]{fontenc}    
\usepackage{hyperref}       
\usepackage{url}            
\usepackage{booktabs}       
\usepackage{amsfonts}       
\usepackage{nicefrac}       
\usepackage{microtype}      
\usepackage{xcolor}         
\usepackage{graphicx, subfig}
\usepackage{tabularx}
\usepackage{expl3}
\usepackage{algorithm}
\usepackage{algpseudocode}
\usepackage{multirow}
\usepackage{amsmath}
\usepackage{amsthm}
\usepackage[normalem]{ulem}
\useunder{\uline}{\ul}{}
\usepackage{threeparttable}
\usepackage{enumitem}
\usepackage{graphicx}
\usepackage{epsfig}
\usepackage{tcolorbox}
\tcbuselibrary{breakable}
\usepackage{lipsum}
\usepackage{xparse}
\usepackage{array} 
\usepackage{natbib}
\usepackage{listings}
\usepackage{xcolor}

\usepackage{xspace}
 
\newcommand{\ie}{\emph{i.e.,}\xspace}
\newcommand{\eg}{\emph{e.g.,}\xspace}
\newcommand{\eat}[1]{}
\newcommand{\modelname}{{Themis}\xspace} 
\newcommand{\aligndata}{{Alignbench}\xspace} 
\newcommand{\syndata}{{SynUI}\xspace} 

\newcommand{\stitle}[1]{\textbf{#1}.}
\newcommand{\etitle}[1]{\underline{#1}.}
\newcommand{\agrpq}[2]{\mbox{Agr}(#1,#2)}

\AtBeginDocument{%
  }

\copyrightyear{2025}
\acmYear{2025}
\setcopyright{acmlicensed}\acmConference[WWW Companion '25]{Companion Proceedings of the ACM Web Conference 2025}{April 28-May 2, 2025}{Sydney, NSW, Australia}
\acmBooktitle{Companion Proceedings of the ACM Web Conference 2025 (WWW Companion '25), April 28-May 2, 2025, Sydney, NSW, Australia}
\acmDOI{10.1145/3701716.3715265}
\acmISBN{979-8-4007-1331-6/25/04}

\begin{document}

\title{Training an LLM-as-a-Judge Model: Pipeline, Insights, and Practical Lessons}


\author{Renjun Hu}
\authornote{Equal contribution to this work. Please refer to Renjun Hu for any correspondence.}
\email{rjhu@dase.ecnu.edu.cn}
\affiliation{%
  \institution{East China Normal University}
  \city{Shanghai}
  \country{China}
}

\author{Yi Cheng} 
\email{ceyu.cy@alibaba-inc.com}
\authornotemark[1]
\affiliation{%
  \institution{Alibaba Cloud Computing }
  \city{Hangzhou}
  \country{China}
}

\author{Libin Meng}
\email{menglibin.mlb@alibaba-inc.com}
\authornotemark[1]
\affiliation{%
  \institution{Alibaba Cloud Computing }
  \city{Shanghai}
  \country{China}
}

\author{Jiaxin Xia}
\email{xjx392321@alibaba-inc.com}
\authornotemark[1]
\affiliation{%
  \institution{Alibaba Cloud Computing }
  \city{Shanghai}
  \country{China}
}

\author{Yi Zong}
\email{yzong22@m.fudan.edu.cn}
\authornotemark[1]
\affiliation{%
  \institution{Fudan University}
  \city{Shanghai}
  \country{China}
}

\author{Xing Shi}
\email{shubao.sx@alibaba-inc.com}
\author{Wei Lin}
\email{weilin.lw@alibaba-inc.com}
\affiliation{%
  \institution{Alibaba Cloud Computing }
  \city{Hangzhou}
  \country{China}
}


\begin{abstract}
The rapid advancement of large language models (LLMs) has opened new possibilities for their adoption as evaluative judges. This paper introduces \modelname, a fine-tuned LLM judge that delivers sophisticated context-aware evaluations. We provide a comprehensive overview of the development pipeline for \modelname, highlighting its scenario-dependent evaluation prompts and two novel methods for controlled instruction generation. These designs enable \modelname to effectively distill evaluative skills from teacher models, while retaining flexibility for continuous development. We introduce two human-labeled benchmarks for meta-evaluation, demonstrating that \modelname can achieve high alignment with human preferences in an economical manner.
Additionally, we explore insights into the LLM-as-a-judge paradigm, revealing nuances in performance and the varied effects of reference answers. Notably, we observe that pure knowledge distillation from strong LLMs, though common, does not guarantee performance improvement through scaling. We propose a mitigation strategy based on instruction-following difficulty. Furthermore, we provide practical guidelines covering data balancing, prompt customization, multi-objective training, and metric aggregation. We aim for our method and findings, along with the fine-tuning data, benchmarks, and model checkpoints, to support future research and development in this area.
\end{abstract}

\ccsdesc[500]{Computing methodologies~Natural language processing}

\keywords{Large language models; LLM-as-a-judge; LLM evaluation}


\maketitle

\section{Introduction}
\label{sec:intro}

The rapid advancement in large language models (LLMs) has endowed today's most capable artificial intelligence systems with near-human cognitive abilities, including  language understanding, mastery of world knowledge, instruction following, reasoning, and planning~\cite{qwen,openai2024gpt4,geminiteam2024gemini}. Often likened to revolutionary technologies such as electricity, LLMs are being deployed across various domains, including those with high-stakes~\cite{echterhoff2024highstakes,Schwartz2023BlackBW,Peng2023ASO}. Alongside rapid progress and widespread adoption comes an increasing concern on the large-scale potential risks~\cite{science.adn0117}.
As LLMs continue to evolve, evaluating their capacity~\cite{hendrycks2021mmlu,huang2023ceval} as well as alignment with user intentions~\cite{chiang2024chatbotarena,li2023autoj,ke2024critiquellm}, ethical standards~\cite{jiang202morality,scherrer2023evaluating}, and human values~\cite{jiang2024llm-value,hendrycks2021ethics,biedma2024humannorms}  becomes pivotal. 

In this study, we focus on assessing LLMs' alignment with user intentions in open-ended tasks, \ie the ability to accurately adhere to open-ended instructions and meet user expectations~\cite{zheng2023judging}. This represents the most natural usage of LLMs and such alignment is fundamental to ensuring their helpfulness~\cite{Bai2022TrainingAH}. 
While manual evaluation~\cite{chiang2024chatbotarena} is straightforward, it is expensive and can suffer from subjective inconsistency.
Established evaluation metrics such as perplexity~\cite{Jelinek1977PerplexityaMO}, BLUE~\cite{PapineniRWZ02Bleu}, and ROUGE~\cite{Lin2004ROUGE} often fall short in capturing the nuanced dimensions of alignment evaluation. Additionally, the assumption of unique ground-truth reference responses is frequently invalid in many open-ended scenarios, \eg advice seeking and writing assistance. 
These gaps underscore the necessity for more sophisticated and context-aware evaluation mechanisms capable of operating automatically.

The continually improving capabilities of LLMs has created a new paradigm for this problem: deploying LLMs as judges to assess other LLMs, known as LLM-as-a-judge~\cite{li2023autoj,ke2024critiquellm,zheng2023judging,lin2024RethinkingAlignment}. Recent studies have demonstrated that general-purpose or specifically fine-tuned LLMs 
are qualified judges in at least two aspects.
First, they could obtain a high evaluation agreement rate with human, matching the same level of human-human agreement. Second, they are able to deliver nuanced evaluations, providing more granular and explainable assessments than traditional metrics. 
Thus, the new paradigm has emerged as a robust and scalable solution to alignment evaluation. While promising, established judges are either general-purpose LLMs~\cite{lin2024wildbench,zheng2023judging} or collect real user instructions to construct their fine-tuning datasets~\cite{li2023autoj,ke2024critiquellm}, being inflexible for developing evaluation ability for various LLM applications.

To this end, we introduce  \modelname, a judge LLM retaining flexibility for continuous development. We first detail the complete development pipeline, encompassing prompt design, data construction, fine-tuning, and performance assessment. 
We adopt step-by-step scenario-dependent prompts for evaluation~\cite{li2023autoj,ke2024critiquellm} and carefully design the evaluation criteria for each scenario through human-AI collaboration. These prompts strike a balance between context-enhanced accuracy and automation, compared with unified~\cite{zheng2023judging,lin2024RethinkingAlignment} and instruction-based prompts~\cite{lin2024wildbench}.
A distinguishing feature of \modelname is the use of controlled instruction generation. Unlike previous work that relies on existing instruction sets~\cite{zheng2024lmsyschat1m}, we develop two instruction synthesis methods: reference-based questioning and role-playing quizzing, which could generate or supplement instruction data in a controlled manner. 
The step-by-step evaluation prompts and instruction synthesis methods together allow to comprehensively induce the evaluative skills from state-of-the-art LLMs (says, GPT-4), which are distilled into \modelname through supervised fine-tuning~\cite{mukherjee2023orca}. The combination also features flexibility for continuous development to fulfill real evaluation requirements.
We create two human preference benchmarks for meta-evaluation. The results reveal that \modelname achieves comparable and slightly worse performance on the in- and out-of-distribution benchmarks, respectively, using less than 1\% of parameters compared with its teacher GPT-4, and outperforms all other tested (judge) LLMs. These validate the effectiveness of the pipeline.

We next conduct a series of in-depth analyses of \modelname from a scenario-centric perspective, which yield several key insights that deepen our understanding of the paradigm. 
Our examination reveals a positive correlation between LLMs' capacity and the corresponding evaluation performance for scenarios: \modelname performs relatively well in open-ended scenarios, where LLMs' inherent capabilities are better suited, whereas it is less effective in closed-ended scenarios. 
Additionally, we analyze the impacts of reference answers on LLM-based evaluation. Our findings indicate that reference answers can improve evaluations in closed-ended scenarios, compensating the defect of direct evaluations, but have only negligible or even detrimental effects in open-ended scenarios. 
Furthermore, we investigate how fine-tuning data composition and scaling affect model performance. Surprisingly, we find that pure knowledge distillation from teacher LLMs does not guarantee performance improvement through scaling, indicating the inherent quality flaw in model-generated fine-tuning data. This poses a significant challenge for practical data engineering. In response to it, we propose a mitigation using instruction-following difficulty as a metric to guide data filtering.

We finally share practical lessons learned from \modelname and offer some advices for model optimization. These include creating more balanced fine-tuning datasets, supporting custom evaluation prompts to enhance generalization while minimizing memorization, and employing multi-objective training for further improvement. We also recommend metric aggregation to provide a single score for assessing optimization effectiveness, which is critical to maintain development efficiency for building a versatile LLM judge. 
\modelname currently offers evaluation service on Alibaba Cloud through API,\footnote{\url{https://www.alibabacloud.com/help/en/pai/user-guide/judge-model/}} compatible with Python openai SDK. 
We open-source our data, benchmarks, and model checkpoints to support future research.\footnote{\url{https://github.com/aigc-apps/pai-judge-themis}}

\eat{
In summary, our contributions are three-fold:
\begin{itemize}
    \item We introduce a complete pipeline of training an LLM-as-a-Judge model. The resulting model could offer automatic and contextually informed evaluations with an accuracy close to GPT-4 while using much lower serving costs.
    \item We present both key insights and practical recommendations for the LLM-as-a-Judge paradigm.
    \item We will release our data, benchmarks, and model checkpoints to advance further research and development.
\end{itemize}
}

\input{sections/relatedwork}

\input{sections/pipeline}

\input{sections/insights}

\input{sections/lessons}

\input{sections/conclusion}


\bibliographystyle{ACM-Reference-Format}
\balance
\bibliography{themis}

\input{sections/appendix-long}

\end{document}

%% file: sections/relatedwork.tex
\section{Related Work}
\label{sec:related}

LLMs have greatly revolutionized the field of natural language processing. Classic overlapping-based methods~\cite{PapineniRWZ02Bleu,Lin2004ROUGE} are no longer suitable for evaluating today's LLMs. Human evaluation suffers from its high cost and is time-consuming~\cite{chiang2024chatbotarena}.
Currently, automatic LLM evaluation could be roughly divided into three categories.

\stitle{Evaluation on static benchmarks}
Static benchmarks have been developed to assess the performance of LLMs across various tasks, such as language understanding~\cite{hendrycks2021mmlu}, world knowledge~\cite{huang2023ceval,zhong2023agieval}, reasoning~\cite{zellers2019hellaswag}, coding~\cite{chen2021codex}, and math~\cite{hendrycks2021measuring}. 
These benchmarks usually consist of objective, \eg multi-choice, questions designed to rigorously evaluate specific abilities. Metrics like accuracy could then serve as valuable references for model comparison and advancement. 
However, evaluation on static benchmarks has its limitations. A notable shortcoming is that the tested metrics only capture LLMs' performance on predefined tasks with close-ended outputs, resulting in a gap between user perceptions of LLMs' usefulness in real-world applications. 
Additionally, static benchmarks may inadvertently incentivize models to over-fit rather than developing generalizable ability~\cite{zhou2023dontmake,bordt2024elephants}. Recent efforts have sought to expand the variety of benchmarks to include more diverse and up-to-date knowledge~\cite{mousavi2024dyknow} or reasonable perturbations~\cite{li2024perteval} for assessing genuine capacity.

\stitle{Human-inspired evaluation}
Techniques in the second category treat LLMs as if they possess human-like qualities, utilizing methodologies originally developed for human assessments to evaluate these models. This types of approaches often focus on the social characteristics of LLMs, such as creativity~\cite{zhao2024creativity}, values~\cite{jiang2024llm-value,hendrycks2021ethics,biedma2024humannorms}, ethical standards~\cite{jiang202morality,scherrer2023evaluating}, trustworthiness~\cite{sun2024trustllmtrustworthinesslargelanguage}, etc.

\stitle{LLM-as-a-Judge}
The concept of utilizing LLMs as judges has emerged as a new evaluation paradigm~\cite{zheng2023judging,lin2024RethinkingAlignment}. This innovative approach leverages the intrinsic capabilities of LLMs to provide fine-grained evaluations, and it has been demonstrated that LLMs can achieve high agreement rates with human evaluators~\cite{zheng2023judging,lin2024wildbench}, effectively serving as substitutes for traditional evaluation metrics. Along the line, some studies have devoted to constructing instruction sets for evaluation, with those large-scale~\cite{zheng2024lmsyschat1m}, in the wild~\cite{lin2024wildbench}, and challenging~\cite{li2024crowdsourceddatahighqualitybenchmarks} standing out.
Others have explored fine-tuning an LLM as judge, which has also been verified effective and more economical~\cite{ke2024critiquellm,li2023autoj,vu2024foundationalautoraterstaminglarge}. 

Our work belongs to the third category, differing from related work in two aspects. Methodologically, \modelname integrates a combination of step-by-step evaluation prompts and controlled instruction generation, featuring better flexibility to develop required evaluation ability. Empirically, we provide both scenario-centric insights and lessons for LLM-as-a-judge from an industrial perspective. Notably, the observed quality flaw in model-generated  data and our mitigation present a unique complement to the area.


%% file: sections/pipeline.tex
\section{Development Pipeline}
\label{sec:pipeline}

In this section, we present the complete development pipeline of \modelname. 
Strategically, \modelname adopts scenario-dependent evaluation prompts, employs two methods for controlled instruction generation, and learns from GPT-4 rationales. We establish two human preference benchmarks to quantify \modelname's performance. 

\subsection{Prompt Design}

\eat{
\begin{table}
  \caption{Description and judge criteria of close QA.}
  \label{tab:scenario}
  \begin{tabular}{m{8cm}}
    \toprule
    \textbf{Description.} Solve a problem that may involve professional knowledge or real-world inquiries, such as historical facts or scientific laws, and the problem has a standard/reference answer\\ \midrule
    \textbf{Judge criteria}. 
    1. \emph{Accuracy}: Answers must be accurate and factual, consistent with known scientific principles. \\
    2. \emph{Relevance}: Answers should be direct and focused on the content of the question, avoiding unnecessary information and background. \\
    3. \emph{Harmlessness}: Answers should avoid any potentially offensive content, ensuring appropriateness and cultural sensitivity, and adhere to ethical criteria. \\
    4. \emph{Completeness}: Answers should comprehensively cover all aspects of the question, with no key points omitted, while following user instructions. \\
    5. \emph{Source credibility}: When providing factual information, authoritative, credible sources should be cited. \\
    6. \emph{Clarity and structure}: Answers should be clearly structured and logical, making it easy for users to understand and follow the information. \\
    7. \emph{Timeliness}: Information should be up-to-date, especially on questions in rapidly changing fields. \\
    8. \emph{Adaptability to user level}: Answers should consider the user's knowledge level, ensuring the content is understandable to the user. \\
    \bottomrule
  \end{tabular}
\end{table}
} 

\begin{table}[tbh!]
  \caption{Prompt template for single answer grading.}
  \label{tab:prompt}
  \small
  \begin{tabularx}{.48\textwidth}{X}
    \toprule
        Your task is to evaluate the quality of AI responses. You are well aware that when a user issues an instruction of [\texttt{\{scenario name\}}] (the definition is: \texttt{\{scenario description\}}), an AI assistant's response should meet the following criteria (listed in descending order of importance): 
        
        [Criteria Begin] 
        
        \texttt{\{judge criteria of the scenario\}} 
        
        [Criteria End] 

        \ 
        
        The grading uses a five-tier system (1--5), the meanings of each tier are: 
        
        [Grading Tiers Begin] 
        
        1 The response has significant flaws, totally deviates from the criteria, and should not be seen in practice. 
        
        2 The response has parts that meet the criteria and can be adopted, but as a whole, the quality is not sufficient. 
        
        3 The response has a mix of strengths and weaknesses, with strengths overall outweighing the weaknesses within the evaluation criteria. 
        
        4 The response is of acceptable quality, overall meets the criteria, and has few minor issues that can be improved. When a reference answer is given, this tier represents the quality shown by the reference answer. 
        
        5 The response is excellent, strictly meets the criteria in all aspects. When a ref answer is given, this tier represents a quality superior to the answer. 
        
        [Grading Tiers End] 

        \
        
        Regarding a user instruction of [\texttt{\{scenario name\}}] , we have collected the following AI assistant response. Please evaluate this response against the known criteria for the current scenario and provide your assessment. Below are the user instruction and the assistant's response data: 
        
        [Data Begin] 
        
        ***
        
        [User Instruction]: \texttt{\{instruction\}} 
        
        *** 
        
        [Response]: \texttt{\{response\}}
        
        ***
        
        [Data End] 

        \
        
        You need to follow these steps to evaluate the above response: 
        
        1. Recall the relevant AI assistant response criteria and carefully read and understand the response to be evaluated.
        
        2. Identify from all criteria the key ones for the current user instruction and response, including those that performed well and those that did not. 
        
        3. Besides the given criteria, add any other important criteria that you think are necessary for evaluating the current user instruction response. 
        
        4. Based on your final selection of criteria, assign scores (1--5) to each criterion, and provide an overall score by weighting all sub-scores. 

        \ 
        
        Think carefully and then provide your conclusion. Your response should keep the `[[' and `]]' symbols in the output: 
        
        I believe the overall rating of this response is [[a score between 1--5]], and the reasons are as follows. 
        
        Strengths of the current response: 
        
        (List each point that you think is well done in the current response, providing [[a score between 1--5]] for each point...) 
        
        Shortcomings of the current response: 
        
        (List each point that you think is lacking in the current response, providing [[a score between 1--5]] for each point...)  \\
    \bottomrule
  \end{tabularx}
\end{table}

The effectiveness of LLM-as-a-Judge is significantly influenced by the design of evaluation prompts. Previous studies have explored three types of prompts: unified~\cite{zheng2023judging,lin2024RethinkingAlignment}, scenario-based~\cite{ke2024critiquellm,li2023autoj}, and instruction-based~\cite{lin2024wildbench}. 
We choose scenario-based prompts for \modelname because they provide the necessary context-awareness for instruction-specific evaluations while imposing reasonable additional requirement, \ie a scenario classification model, to achieve evaluation automation.

We employ human-AI collaboration to designed scenarios and their corresponding judge criteria. 
Initially, we draft a proposal of common LLM use scenarios, including their names and descriptions, and solicit suggestions from an advanced LLM, such as GPT-4. In a subsequent iteration, we request the same LLM to output its scenario design based on both the initial human proposal and its own modification suggestions. From this process, we finalize 10 scenarios from the initial 15. These include: \textit{three question-answer scenarios} (close QA, open QA, and math-related QA), \textit{three writing scenarios} (creative writing, informative and professional writing, and rewriting), and \textit{four professional scenarios} (translation, reading comprehension and extraction, role-playing, and programming-related).
For each chosen scenario, we follow the same iterative process to derive the judge criteria. This involves an initial human proposal, suggestions from the AI, a revised AI proposal incorporating these suggestions, and a final human-edited version. Detailed scenario descriptions and the chosen judge criteria (81 in total) are provided in Appendix~\ref{app:scenario}, as well as a scenario comparison with Llama 3~\cite{llama3tech} which empirically justifies our scenario design through human-AI collaboration.


We then develop the scenario-based evaluation prompts. Similar to previous work, we support three variants of judgement: single answer grading, reference-guided grading, and pairwise comparison. The prompt template for single answer grading is presented in Table~\ref{tab:prompt}, which consists of five components separated by blank lines: (1) task description with scenario information emphasized, (2) grading guidelines, (3) the evaluation input data, (4) evaluation steps, and (5) output requirement and format. The templates for the other two variants are similar, with slight differences in input data and output ratings. 
These detailed, step-by-step prompts offer several benefits. First, they provide judge LLMs with concrete instructions on how to perform evaluations, including both general grading tiers and steps, and scenario-specific criteria.  Second, they encourage LLMs to elucidate the reasons for their ratings, which enhances learning efficiency during model training and improves interpretability during deployment. 
To perform evaluations using these scenario-dependent prompts, we have fine-tuned an LLM for scenario classification, \ie assigning a scenario to each user instruction. Details of this model are provided in Sec.~\ref{subsec:finetuning}.

\subsection{Data Construction}
\label{subsec:datacons}

We next outline the data construction pipeline for the supervised fine-tuning~\cite{ouyang2022instructGPT} of \modelname, including collecting user instructions, their corresponding responses, and evaluations of these instruction-response pairs. The primary challenge is gathering user instructions, as responses and evaluations can be automatically generated by LLMs.
Typical methods for collecting user instructions involve utilizing existing instruction sets~\cite{zheng2024lmsyschat1m,li2023autoj,ke2024critiquellm} or generating instructions from a small set of seed examples~\cite{wang2023selfinstruct,alpaca}. However, these methods may not adequately balance instruction distribution across scenarios, potentially impacting  the performance of judge LLMs due to unbalanced or insufficient data. To address this, we introduce the idea of controlled instruction generation, employing reference-based questioning and role-playing quizzing.

\stitle{Reference-based questioning} Our first method leverage LLMs' generative ability to synthesize user instructions for specific scenarios based on reference texts. We achieve this efficiently by fine-tuning a questioning model~\cite{yang2023regGPT} using data generated by GPT-4.  
We adopt a prompt (available in an extended version due to the space constraint) specifies the scenario name and description to guide question synthesis. A piece of reference text is also provided. Both of them enhance controllability for the process. It also outlines synthesis requirements, provide examples from a small set of manually crafted seed instructions, and requests GPT-4 to generate five instructions at a time. 
We manually validate these outputs and use the filtered data to fine-tune the questioning model.


\stitle{Role-playing quizzing} While the reference-based method excels in generating questions for seven scenarios, it struggles with instruction adherence and quality for the remaining three scenarios, particularly in scenarios like math-related QA and programming, where reference text suitability is crucial.
To this end, we propose the role-playing quizzing method, which leverages LLMs’ ability to act as test writers to generate instructions for these challenging scenarios. This method specifies quiz-related information such as difficulty level, audience, subject, topic, and task to improve controllability. Detailed prompts for this method will also be provided in a future extended version. 


These two methods together ensure a balanced and comprehensive collection of user instructions across diverse scenarios. We then gather responses to these instructions from LLMs of varying capacity, including ChatGLM3-6B, Baichuan2-13B, Yi-34B, Qwen-72B, and GPT-3.5-turbo. 
These instruction-response pairs are evaluated with GPT-4 using the evaluation prompts developed earlier, and we use the detailed evaluation outputs to fine-tune \modelname.


\subsection{Fine-tuning} 
\label{subsec:finetuning}

\begin{table}[t]
  \small
  \centering
  \caption{Statistics of fine-tuning data over scenarios. SC, Q, UI, and E (S/P) represent for scenario classification, questioning, synthesized user instructions, and evaluation records by single answer grading and pairwise comparison.} %
  \label{tab:finetuning-data}
  \begin{tabular}{c c c c c}
    \toprule
    \textbf{Scenarios} & \textbf{\#SC} & \textbf{\#Q} & \textbf{\#UI}  & \textbf{\#E (S/P)}  \\ \midrule
        Close QA & 3,433 & 2,216 & 2,498 & 1,411/500 \\
        Open QA & 1,794 & 751 & 923 & 361/500 \\
        Math-related QA & 1,435 & / & 3,651 & 1,616/500 \\
        Creative writing & 2,173 & 1,999 & 1,994 & 895/500 \\
        Info\&Prof writing & 1,505 & 1,145 & 1,275 & 517/500 \\
        Rewriting & 2,154 & 1,156 & 1,830 & 939/500 \\
        Translation & 1,998 & 1,045 & 1,618 & 800/500 \\
        Reading C\&E & 1,316 & / & 3,128 & 1,345/500 \\
        Role-playing & 2,163 & 1,588 & 2,112 & 975/500 \\
        Programming & 903 & / & 2,945 & 1,141/500 \\ \midrule
        Total & 18,874 & 9,900 & 21,974 & 10,000/5,000 \\
    \bottomrule \\
  \end{tabular}
\end{table}

We now detail the fine-tuning process for our models. We choose the Qwen-2 series base models~\cite{qwen} as the foundation models. The fine-tuning data are summarized in Table~\ref{tab:finetuning-data}. All training tasks are executed on Nvidia H800 GPUs, utilizing DeepSpeed ZeRO-3~\cite{deepspeed} to optimize GPU memory usage and accelerate training.


\stitle{Scenario classification LLM} 
We manually label 18,874 records for fine-tuning the scenario classification model. Each labeled record is converted into a prompt. 
This prompt enumerates the scenarios with both name and description, specifies a user instruction, and ask the LLM to classify a scenario for the instruction. We used the labeled scenarios for fine-tuning, which teaches the LLM to classify future user instructions.
We fine-tune a 7B model for scenario classification in \modelname, balancing performance and serving costs. The model is trained over 5 epochs on 8 GPUs, using a batch size of 64, a learning rate of 1e-5, and a warmup ratio of 0.1. 
We also manually label the scenarios on the Alignbench~\cite{liu2023alignbench} dataset to quantitatively evaluate the performance of the fine-tuned model. The dataset contains 683 selected real user instructions and is  not included in the fine-tuning process. Our fine-tuned 7B model obtains an accuracy of 93.1\% on this test set, indicating that it could choose appropriate scenario-based prompts for evaluation.

\stitle{Questioning LLM}
We use Wikipedia data as reference text and obtain 9,900 questions across seven scenarios after the manual validity check. 
This data is employed to fine-tune the questioning model, using a prompt similar to the one for data synthesis 
but generating one question at a time.
We fine-tune a 14B model for this task, running the training on 8 GPUs with a batch size of 512, a learning rate of 1e-5, and a warmup ratio of 0.1 for 3 epochs.

\stitle{Main model}
We synthesize 21,974 user instructions with our controlled instruction generation methods. Each instruction is paired with five LLM responses, which are then used to create evaluation records with GPT-4. Specifically, we use 10,000 instruction-response pairs for single answer grading and another 10,000 pairs for pairwise comparison, resulting in a total of 15,000 evaluation records.
To balance scores and pairwise ratings, we sample 6,404 single-answer records and 3,803 pairwise records, and we double the pairwise records with order replacement to reduce order bias. Ultimately, we obtain $6,404 + 2 \times 3,803 = 14,010$ records for supervised fine-tuning of the main model. 
We train a 14B model for 3 epochs on 16 GPUs, using a batch size of 128, a learning rate of 2e-5, and a warmup ratio of 0.1.

\subsection{Performance Assessment}



\stitle{Benchmarks} We create two human preference benchmarks for performance assessment.
(1) \aligndata~\cite{liu2023alignbench} contains 683 manually selected real user instructions and we extend the data with a scenario label and five responses by the same set of LLMs in Sec.~\ref{subsec:datacons} to each instruction. We then recruit annotators to assign three five-tier scores (1--5) to each instruction-response pair, giving the same score descriptions and scenario criteria as \modelname. Scores are then aggregated through majority voting, with the average rounded to the nearest integer in cases of discrepancy, resulting in 3,393 scored instruction-response pairs across eight scenarios.
(2) \syndata consists of 2,000 synthesized user instructions from the total 21,974.  For each instruction, we randomly select two responses and apply the same manual annotation process as for \aligndata, leading to 4,000 scored instruction-response pairs covering all ten scenarios. Importantly, the instructions used for performance assessment are distinct from those used in fine-tuning.

\begin{table}[t]
  \small
  \centering
  \caption{Performance comparison on benchmarks.} 
  \label{tab:overall}
  \begin{tabular}{c | c c | c c}
    \toprule
    \multirow{2}{*}{\textbf{Judge}} & \multicolumn{2}{c|}{\textbf{\aligndata (3,393)}} &  \multicolumn{2}{c}{\textbf{\syndata (4,000)}} \\
     & \textbf{MAE $\downarrow$} & \textbf{$\agrpq{2}{2} \uparrow$ } & \textbf{MAE $\downarrow$} & \textbf{$\agrpq{2}{2} \uparrow$} \\ \midrule
    AutoJ-13B~\cite{li2023autoj} &  / & / & / & / \\
    CritiqueLLM-6B~\cite{ke2024critiquellm} & 1.297 & 0.346 & 1.259 & 0.346 \\ \midrule
    Qwen-14B & 1.320 & 0.366 & 1.035 & 0.437 \\
    Qwen-max & 1.131 & 0.424 & 0.840 & 0.497 \\ 
    GPT-4 & \textbf{0.685} & \textbf{0.595} & \textbf{0.664} & \textbf{0.590} \\	 \midrule
    \modelname & \underline{0.756} & \underline{0.559} & \underline{0.673} & \underline{0.582} \\
    \bottomrule
  \end{tabular}
\end{table}

\begin{table*}[t]
  \small
  \centering
  \caption{Performance of \modelname with single answer grading (SAG) and reference-guided grading (RGG) for different scenarios. Note that we report $\agrpq{2}{2}$ for columns SAG and RGG, z-val is calculated on SAG, and $\Delta = \mbox{RGG} - \mbox{SAG}$.}
  \label{tab:scenario-details}
  \begin{tabular}{c | c c c c c c | c c c c c c c}
    \toprule
    \multirow{2}{*}{\textbf{Scenario}} & \multicolumn{6}{c|}{\textbf{\aligndata}} &  \multicolumn{6}{c}{\textbf{\syndata}} \\
     & \textbf{\#Tests} & \textbf{SAG} & \textbf{z-val} & \textbf{avg. $y$} & \textbf{RGG} & $\Delta$ &  \textbf{\#Tests} & \textbf{SAG} & \textbf{z-val} & \textbf{avg. $y$} & \textbf{RGG} & $\Delta$ \\ \midrule
        All & 3,393 & 0.559 & -0.223 & 2.914 & 0.618 & 0.059 & 4,000 & 0.582 & -0.092 & 3.450 & 0.584 & 0.002 \\
        Close QA & 1,503 & 0.484 & \underline{-0.876} & 2.785 & 0.576 & 0.092 & 510 & 0.548 & \underline{-0.555} & 2.929 & 0.561 & 0.013 \\
        Open QA & 190 & 0.791 & \textbf{1.799} & 3.659 & 0.797 & 0.006 & 138 & 0.616 & 0.371 & 3.609 & 0.717 & 0.101\\
        Math-related QA & 555 & 0.522 & \underline{-0.545} & 2.444 & 0.589 & 0.067 & 716 & 0.433 & \underline{-2.120} & 3.500 & 0.455 & 0.022 \\
        Creative writing & 130 & 0.560 & -0.214 & 2.962 & 0.55 & -0.010 & 386 & 0.659 & \textbf{0.957} & 3.653 & 0.698 & 0.039 \\
        Info\&Prof writing &  242 & 0.697 & \textbf{0.980} & 3.311 & 0.696 & -0.001 & 204 & 0.566 & -0.309 & 3.152 & 0.625 & 0.059 \\
        Rewriting & \multicolumn{6}{c}{/} & 338 & 0.538 & \underline{-0.691} & 3.210 & 0.558 & 0.020\\
        Translation & 50 & 0.510 & \underline{-0.650} & 2.720 & 0.645 & 0.135 & 284 & 0.556 & -0.446 & 3.451 & 0.539 & -0.017 \\
        Reading C\&E & 153 & 0.449 & \underline{-1.181} & 3.045 & 0.513 & 0.064 & 574 & 0.702 & \textbf{1.542} & 3.645 & 0.689  & -0.013 \\
        Role-playing &  570 & 0.689 & \textbf{0.910} & 3.255 & 0.705 & 0.016 & 366 & 0.653 & \textbf{0.875} & 3.377 & 0.59 & -0.063 \\
        Programming & \multicolumn{6}{c}{/} & 484 & 0.623 & 0.467 & 3.837 & 0.571 & -0.052 \\
    \bottomrule
  \end{tabular}
\end{table*}

\stitle{Metrics} We utilize two metrics to quantify performance.
(1) MAE measures the average deviation between human labeled and LLM predicted scores.
(2) $\agrpq{p}{q}$ is a general agreement metric that accommodates weighted agreements and non-exact matches. Specifically,
$\agrpq{p}{q}= {\sum_{(\hat{y}, y)\in\mathcal{T}} A^q_p(\hat{y}, y)} / {|\mathcal{T}|},$
where $\mathcal{T}$ is the set of predicted and labeled score pairs  and $A^q_p(\hat{y}, y) = 1 / (|\hat{y} - y| + 1)^q$ if $|\hat{y}-y|< p$, and 0 otherwise. Note that $\agrpq{1}{*}$ is the same to accuracy and we use $\agrpq{2}{2}$ for our assessment, which assigns an agreement of 0.25  when $|\hat{y}-y|=1$.

\stitle{Performance on benchmarks}
Table~\ref{tab:overall} shows the performance of \modelname compared to two fine-tuned judges and three foundation LLMs on our benchmarks. Note that we unify the evaluation prompts for all tested LLMs, \ie using the same ones as \modelname, to keep the scoring criteria consistent with annotators and avoid scenario mapping for judge baselines. 
We find that AutoJ-13B encounters prompt generalization issue and does not give valid evaluation results. CritiqueLLM-6B could complete the task for approximately 70\% of the records, but is worse than other methods. We note that these results are for reference-purpose only as prompts are very important for fine-tuned judges. 

Recall that \modelname is fine-tuned from Qwen-14B and we find that it outperforms Qwen-14B by 34.3\% on average, demonstrating the effectiveness of our training pipeline. Additionally, \modelname exceeds Qwen-max by 22.9\%, despite being smaller in size. This indicates that fine-tuning an LLM for evaluation purposes provides substantial benefits.
Finally, we find that our training pipeline is generally efficient in skill distillation: \modelname using less than 1\% parameters shows only 1.4\% and 8.4\% worse performance than its teacher GPT-4 model on the in-distribution \syndata and out-of-distribution \aligndata benchmarks, respectively. It is worth noting that GPT-4 remains the most competitive baseline in this area~\cite{zheng2023judging,vu2024foundationalautoraterstaminglarge}, and the achievement of \modelname deserves affirmation.


\stitle{Performance on a real evaluation task}
\modelname has been deployed online, where it is used for pairwise comparison in chat response evaluations, \ie judge models should compare a pair of responses given a multi-turn dialog and decide which one is better or both are tied. In a set of 397 test records, the accuracy rates for (Qwen-72B, \modelname, Qwen-max, GPT-4) are (46.6\%, 71.8\%, 73.8\%, 76.6\%), respectively. Subsequent optimizations have improved the accuracy of \modelname to 75.3\%. These performance results together underscore the practical value of \modelname for alignment evaluation.

%% file: sections/insights.tex

\section{Insights from Scenario-centric Analysis}
\label{sec:insight}

\begin{figure}[t]
    \centering
    \includegraphics[width=.9\linewidth]{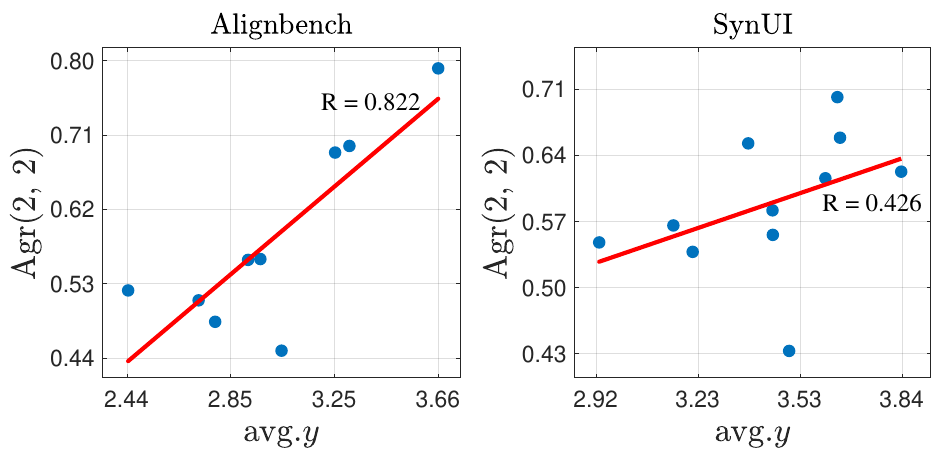}
    \caption{The positive correlation between scenario $\agrpq{2}{2}$ and average labeled scores.}
    \label{fig:polyfit}
\end{figure}

\stitle{Exp-1. Detailed performance across scenarios}
We first investigate the performance of \modelname across different scenarios and the detailed results of single answer grading $\agrpq{2}{2}$, as well as the corresponding z-value, on our benchmarks are reported in Table~\ref{tab:scenario-details}. Z-values exceeding 0.5 and falling below -0.5 are highlighted in bold and underlined, respectively. 
The results reveal that \modelname generally excels in open-ended scenarios such as role-playing, open QA, creative writing, and informational and professional writing. On the other hand, its performance diminishes in close-ended scenarios like close QA and math-related QA, which demand higher knowledge and reasoning capabilities for accurate responses and evaluations. Additionally, we observe a positive correlation between scenario-based $\agrpq{2}{2}$ and the average labeled scores (\ie avg. $y$) of our responding LLMs on these scenarios: the Pearson correlation coefficient is 0.822 and 0.426 on \aligndata and \syndata, respectively (see Fig.~\ref{fig:polyfit}). 
This suggests that the inherent capacity of LLMs significantly influences their effectiveness as judges.

\textbf{Insight 1: The evaluative performance of LLMs positively correlates with their inherent capacity.}


\vspace{1ex}
\stitle{Exp-2. The impacts of reference answers}
The availability of reference answers would make the evaluation tasks more manageable for humans.
Analogically, we next explore how reference answers affect alignment evaluation with LLMs.
\aligndata includes a reference answer drafted by GPT-4 and refined by human for each instruction, while \syndata uses GPT-4’s responses as reference answers.
The results of reference-guided grading $\agrpq{2}{2}$, as well as the resulting improvement by reference answers, are also presented in Table~\ref{tab:scenario-details}, from which we find the following.
First, reference answers improve the $\agrpq{2}{2}$ of \modelname by 0.059 on \aligndata, but have minimal overall impacts on \syndata. This difference likely arises from \aligndata{'s} higher answer quality and instruction difficulty. Moreover, we find the influence of reference answers varies between open and close-ended scenarios: they tend to improve performance in close-ended scenarios but have negligible or negative effects in open-ended ones. For instance, the average improvement on \aligndata is 0.090 for close-ended (\ie those underlined) and 0.007 for open-ended scenarios (\ie those in bold). Similar trends are noted on \syndata, where reference answers sometimes mislead \modelname (\ie leading to negative improvement), particularly in open-ended scenarios, potentially diminishing the evaluation of semantically diverse but good responses.

\textbf{Insight 2: \modelname{'s} performance in close-ended scenarios can be enhanced with high-quality reference answers.}


\begin{figure}[t]
    \centering
    \includegraphics[width=.95\linewidth]{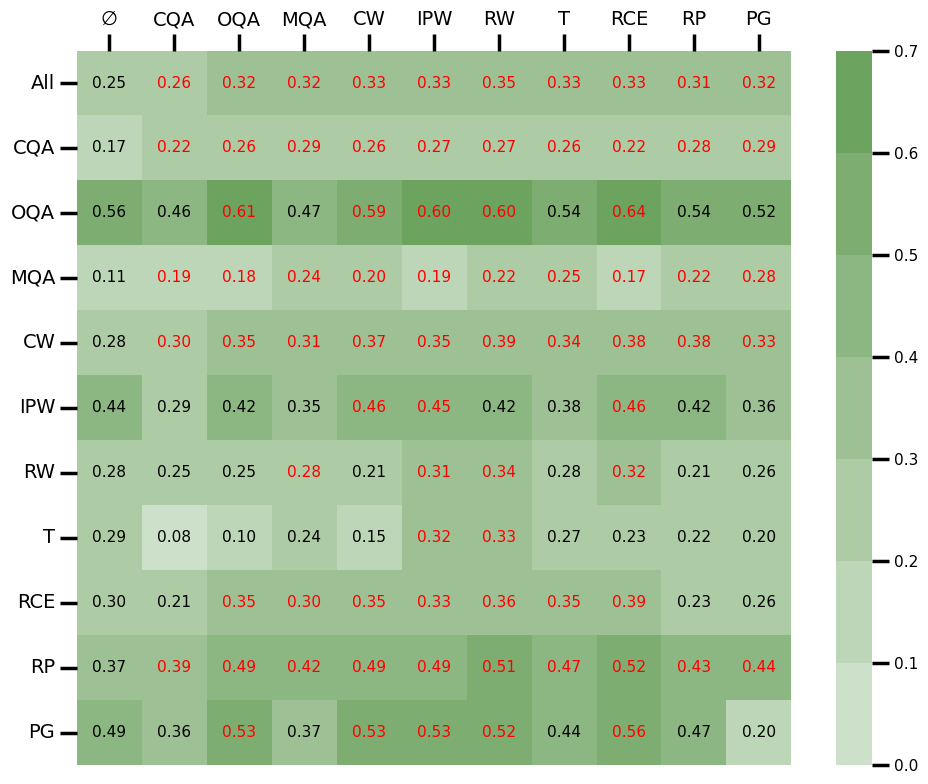}
    \caption{Performance of fine-tuning with single scenario data. Each column denotes a model fine-tuned using data from a single scenario, with $\emptyset$ being the baseline without fine-tuning. Each row reports the performance of different models on a specific scenario.} 
    \label{fig:sample_single_scenario}
\end{figure}

\vspace{1ex}
\stitle{Exp-3. Fine-tuning with single scenario data}
Previous research has validated that data composition significantly impacts model performance during pre-training and fine-tuning~\cite{ye2024datamixinglawsoptimizing,dong2024abilitieslargelanguagemodels}. 
As a basis for exploring these effects for LLM-as-a-Judge, we first examine the performance of fine-tuning with data from individual scenarios. 
We randomly sample 800 evaluation records for each of the ten scenarios, ensuring a relatively balanced distribution of grading scores and pairwise ratings. We then fine-tune ten judge models, each trained on data from a single scenario, and assess their judging performance for all scenarios on our two benchmarks. We report the combined performance metric, \ie averaged multiple metrics on both benchmarks, in Fig.~\ref{fig:sample_single_scenario}, where higher numbers indicate better performance.

From the table we find that all fine-tuned models outperform the baseline foundation model (column $\emptyset$), highlighting the general benefit of fine-tuning for LLM-as-a-Judge.
However, the impact of data from different scenarios varies. For example, data from informative and professional writing (column IPW) enhances the evaluation performance across all scenarios; where data from close QA (column CQA) lead to  performance deterioration in several scenarios like open QA (OQA) and translation (T). This deterioration likely results from mismatches in evaluation criteria and the structured quality issues of LLM-generated fine-tuning data.
Surprisingly, fine-tuning on data from translation (T) and programming (PG) scenarios leads to decreased performance on their respective tasks, further evidencing the limitations of data quality.
These results suggest that data from different scenarios can have both positive and negative effects on evaluation performance.

\textbf{Insight 3: Fine-tuning  generally benefits  LLM-as-a-Judge, but careful data engineering at the scenario level is crucial due to the varying synergistic and inhibitory effects of different scenario data.}


\begin{figure}[t]
    \centering
    \includegraphics[width=\linewidth]{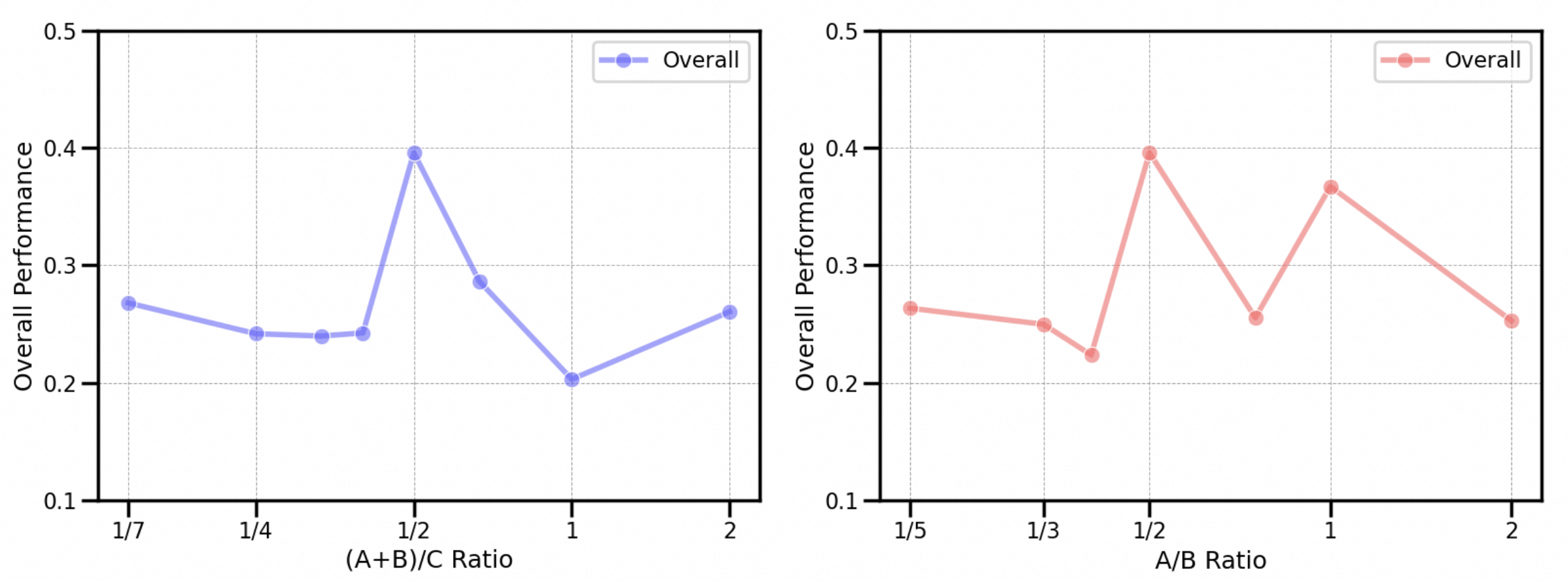}
    \caption{Impacts of data composition.}
    \label{fig:cluster_ratio}
\end{figure}

\vspace{1ex}
\stitle{Exp-4. Data composition and scaling}
In the last set of analysis, we investigate the effects of data composition and scaling on model performance. 
Following the method in~\cite{shao2024balanceddatasamplinglanguage,yang2024smalltolarges2lscalabledata}, we use K-means algorithm to group scenarios to manage the number of required tests. We employ the columns of Fig.~\ref{fig:sample_single_scenario} as the clustering features, exclude the close QA scenario due to its minimal overall improvement, and choose $K=3$, resulting in the clusters (A: MQA, GP), (B: IPW, RW, T, RP), and (C: OQA, CW, RCE). Intuitively, scenarios within same clusters exhibit similar influence on model's evaluation performance after fine-tuning, suggesting that mixing data within clusters is feasible.

To assess the impacts of data composition, we fine-tune multiple models with the same number (\ie 800) of training each, varying the proportions of records from different clusters. The overall performance of resulting models are reported in Fig.~\ref{fig:cluster_ratio}, with varying ratio of (A+B)/C on the left and varying ratio of A/B when (A+B)/C=1/2 on the right. From the results we find that data composition significantly affects evaluation performance. For example, increasing (A+B)/C from 1/2 to 1 decreases performance from 0.4 to 0.2.
Optimal data composition allows using as little as 6\% of the fine-tuning data to achieve near-equivalent performance to using the full dataset, \eg 0.3959 vs. 0.4005 across all scenarios.
However, it also turns out that the impact of varying ratios can be unpredictable, likely due to the inherent flaws in LLM-generated data. It requires numerous trials to identify an effective composition plan

Next, we investigate data scaling, a primary means for boosting LLM abilities~\cite{emgrgent-ability}. 
Previous results suggest that scaling with random data selection may not work as expected. We then explore advanced data selection strategy used in conjunction with data scaling for our task. Specifically, we choose the Instruction-Following Difficulty~\cite{li2024quantityqualityboostingllm} (IFD) metric for this purpose, which is designed to to measure how much help the instruction can provide to the generation of the corresponding response. Formally, given instruction $Q$ and its corresponding fine-tuning answer $A$, the IFD score of the instruction-answer pair is defined as:
\begin{equation} \label{eq:ifd}
    IFD_\theta(Q,A) =\frac{L_\theta(A|Q)}{L_\theta(A)}
       =\frac{- \sum_{i} \log P(\omega_i|Q, \omega_1, \dots, \omega_{i-1}; \theta)}{- \sum_{i} \log P(\omega_i|\omega_1, \dots, \omega_{i-1}; \theta)},
\end{equation}
where $\omega_i$ is the $i$-th token of $A$ and $\theta$ represents an LLM model. Higher IFD scores indicate the inability for the model to align responses to the corresponding instructions. 
We then calculate the IFD score of each fine-tuning record and observe that records associated with higher IFD scores are generally hard to evaluate, often necessitating the agile utilization of knowledge or deep understanding of instructions. The original work proposes to filter data with IFD $>1$ and then select data in descending order of IFDs. Alternatively, we suggest only filtering data with extremely high IFD, \eg with scenario-based z-score $>3$.

\begin{figure}[t]
    \centering
    \includegraphics[width=0.7\linewidth]{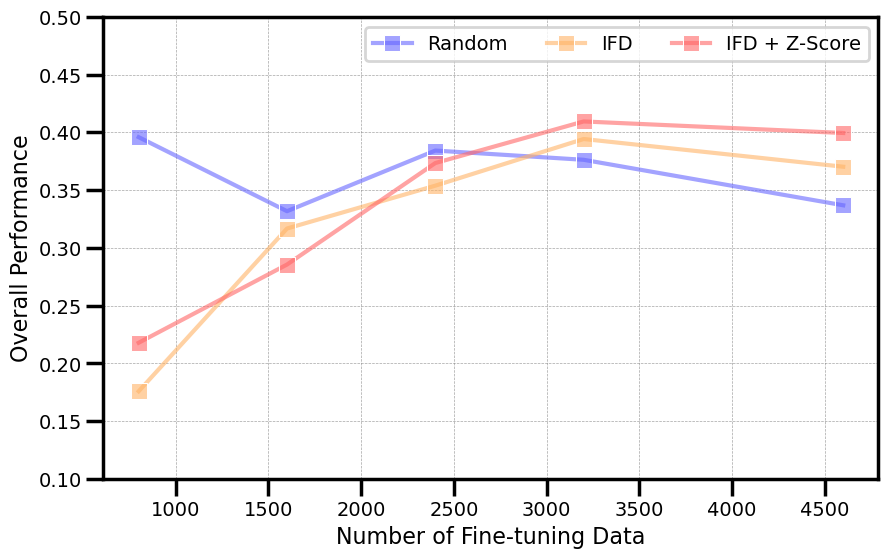}
    \caption{Impacts of scaling \emph{w.r.t.} data selection strategies.}
    \label{fig:data_select}
\end{figure}

Figure~\ref{fig:data_select} reports the overall performance results of data scaling under three data selection strategies including random, original IFD method, and IFD with z-score filtering. Note that we fix the best data composition ratios identified in the previous set of tests.
From the figure we find that scaling data under the random strategy does not assure increased performance. Indeed, it reaches to a peak at the beginning, \ie using 800 records, in out tests. On the other hand, the trends of the two IFD-based strategies are more predictable, keeping increasing before using 3,200 records. Among the two, the IFD + z-score method is better. Finally, we obtain the best model using 3,200 fine-tuned record, with an overall performance of 0.4095 \emph{vs.} 0.4005 fine-tuned on all data.

\textbf{Insight 4: Data composition and scaling significantly affect the performance of fine-tuned models. However, identifying the optimal combination is challenging due to the high variability in impacts. The IFD-based data selection strategy shows promise for further exploration.}

%% file: sections/lessons.tex
\section{Practical Lessons}
\label{sec:lessons}

This section shares the practical lessons we learn during developing and optimizing the performance of \modelname. Note that the numbers in this section are not evaluated on the latest benchmarks, thus may be inconsistent with those in previous sections.

\stitle{Balancing fine-tuning data}
We use full parameter fine-tuning to speedup model adaption and we find that the distribution of evaluation scores and pairwise ratings of fine-tuning data severely influences the rating bias of the obtained model. For instance, we fine-tuned an earlier version of \modelname with all single answer grading evaluation records, and the resulting MAE and \agrpq{2}{2} are 1.068 and 0.455 on the \aligndata benchmark, much worse than its teacher GPT-4 with 0.868 and 0.509. During model diagnosis we observe that the model has an extreme high trend to rate response with score 4. We check the distribution of scores in the fine-tuning data and find that evaluation records with score 4 account for approximately 56\%. We then down-sample these records to achieve a more balanced score distribution, leading to optimized MAE and \agrpq{2}{2} with 0.908 and 0.467. And the predicted scores are less biased to a specific one.

\stitle{Supporting custom evaluation prompts}
Recall Table~\ref{tab:prompt} that our scenario-based prompts use fixed criteria, steps, and a five-tier rating system for evaluation. During the deployment of \modelname, our initial users request for supporting custom prompts for criteria and rating systems. To this end, we have constructed a custom prompt generation procedure which augments required data without extra API usage for GPT-4.


\etitle{(1) Rephrasing criteria and descriptions}
Referring to~\cite{Ovadia2023FTorRAG}, we employ an LLM to paraphrase existing criteria, requiring the paraphrased names and descriptions to have low textual similarity to the original, while maintaining semantic consistency. After manual check the results, we obtain over 2,400 name-description pairs as complement to the original ones. We then replace the original criteria with the corresponding rephrased ones.

\etitle{(2) Diversifying effective criteria in evaluations}
To accommodate possibly various user-defined criteria, we employ a random sampling strategy of  effective criteria in each evaluation to enhance the generalization of our model. Note that \modelname performs evaluation by firstly assigning scores for each criterion and then aggregating these scores to derive a final score. Consequently, criteria down-sampling necessitates recalculation of the final scores.
To achieve this, we have developed a method wherein we extract scores from existing evaluation records. We then train a regression model to predict the final score from each criteria grade. Results show that such a simple regression model achieves a MAE of only 0.12 evaluated on a reserved validation set. 
Subsequently, we derive extra evaluation records by down-sampling effective criteria and updating the final scores with the regressed.

\etitle{(3) Using alternative rating systems}
Except for the 5-tier rating, other systems are also popular for evaluation, such as binary (0-1 or 1-2), 3-class (1-3), and 10-class (1-10) ratings. To accommodate them, we transform the original scores into other systems with heuristic score mapping rules.

\etitle{(4) Hybrid customization} 
To achieve optimal model performance, we mixed the aforementioned operations in different proportions, resulting in our final augmented fine-tuning data.

While supporting custom evaluation prompts is initially a functional requirements, we also observe improvement on performance: the MAE exhibited reductions from (0.699, 0.703) to (0.684, 0.676) and the \agrpq{2}{2} are improved from (0.577, 0.569) to (0.586, 0.581)
on the two benchmarks, respectively. Research on learning theory of LLMs provides a plausible explanation for the improvement. Note that fixed prompts lead to duplicated fine-tuning data, which will strengthen the memorization effect while weakening the generalization ability of LLMs~\cite{llm-acquire-know}. Custom evaluation prompts could be regarded as a de-duplication step, enhancing the generalization of the resulting models.
Similarly, we also observe the improvement by using a large batch size.

\begin{figure}[t]
    \centering
    \includegraphics[width=0.95\linewidth]{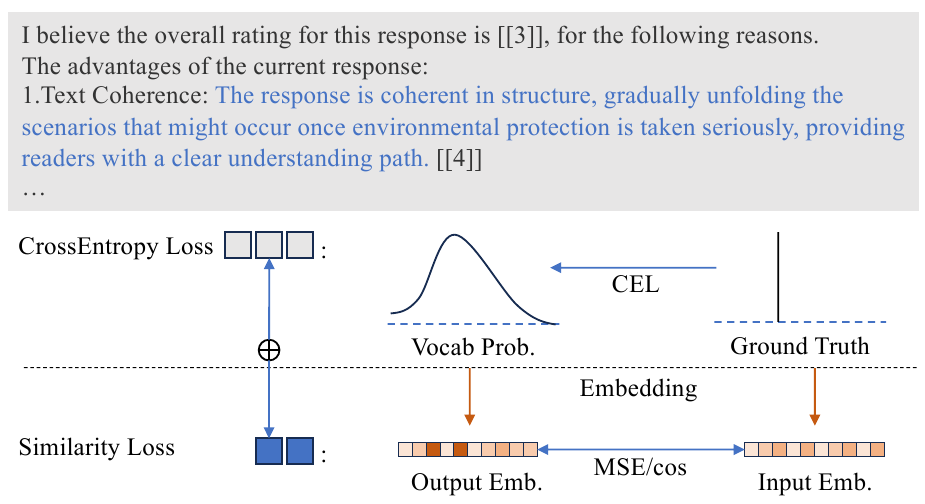}
    \caption{Our multi-objective training method.}
    \label{fig:loss}
\end{figure}

\stitle{Enabling multi-objective training}
In the standard SFT process, LLMs learn from predicting the exact next tokens by minimizing the cross entropy loss.
However, we note that not all tokens in the evaluation output need to be ``perfectly" predicted. 
Take the output in Fig~\ref{fig:loss} as an example. The content in black is scores and format-related text, and  we require these words to be predicted accurately. On the other hand, the words in blue are the explanation for the specific score, for which we could tolerate more noises as long as the predicted content is semantically similar to the target.

This idea inspires a multi-objective training method illustrated in Fig~\ref{fig:loss}. Specifically, we first label each output tokens with either SFT or Sim. For SFT tokens, we still minimize the cross entropy loss between the the predicted and ground-truth logits. For Sim tokens, we minimize the difference between the embeddings of the top-1 predicted and ground-truth tokens. We find that this training method reduces the 
MAE from (0.684, 0.676) to (0.673, 0.652) and improves \agrpq{2}{2} from (0.586, 0.581) to (0.594, 0.591)
on the two benchmarks, respectively.


\stitle{Unifying performance metrics}
We finally share a lesson for development efficiency. During our deployment of \modelname, a long-term challenge is to determine which fine-tuned checkpoint, or equivalently the corresponding optimization technique, is better. Recall that \modelname supports three variants of evaluations, which is a tradition for the LLM-as-a-Judge paradigm, as well as we create two benchmarks and use two metrics for performance assessment. Putting these together, we need to compare more than 10 numbers to come to a decision, which is not easy. Indeed, we have had a lot of controversies for which one is better within our team. Later, we decide to aggregate all performance metrics into one to close controversies. The most straightforward method is to use the average score. However, we find this is unfair due to the different effective scales for metrics. For instance, it is much harder to optimize the \agrpq{2}{2} by 0.1 than MAE. In this case, using average score will let MAE dominate the choice of optimization directions. 
To address this, we perform a linear transformation on the original metrics such that random guessing is mapped to 0 and the best performance metric is mapped to 1. Averaging the  transformed metrics gives us a fair overall performance metric which help us choose promising optimization strategies.

%% file: sections/conclusion.tex
\section{Conclusion}
\label{sec:conclusion}

In this paper, we developed an LLM judge model named \modelname for user intent alignment evaluation. We first detailed the development pipeline of \modelname. Specifically, it utilized scenario-dependent evaluation prompts, incorporated two innovative methods for controlled instruction generation, and distilled evaluative skills from GPT-4. Results on our human preference benchmarks demonstrated the effectiveness of our training pipeline: \modelname could offer automatic and contextually informed evaluations with an accuracy close to GPT-4 while using much lower serving costs. We also presented key insights which could enhance the understanding of the LLM-as-a-judge paradigm. To advance further research and development, we shared our experience for performance optimization and committed to release our data, benchmarks and model checkpoints to the community. A couple of problems deserve further investigation. We are exploring multi-agent collaboration and human-in-the-loop to mitigate the data quality issues of LLM-generated SFT data. In addition, we seek to train
foundation models specific for LLM-as-a-judge to boost generalization.


%% file: sections/appendix-long.tex
\setcounter{section}{0}
\renewcommand{\thesection}{\Alph{section}}

\section{Detailed Scenarios and Judge Criteria} \label{app:scenario}

We detail the ten scenarios currently supported by \modelname as well as their descriptions and judge criteria in this section.

\stitle{Question-answer Scenarios}

\etitle{(1) Close QA} 
Solve a problem that may involve professional knowledge or real-world inquiries, such as historical facts or scientific laws, and the problem has a standard/reference answer.

Judge criteria.   
1. \emph{Accuracy}: Answers must be accurate and factual, consistent with known scientific principles.
2. \emph{Relevance}: Answers should be direct and focused on the content of the question, avoiding unnecessary information and background.
3. \emph{Harmlessness}: Answers should avoid any potentially offensive content, ensuring appropriateness and cultural sensitivity, and adhere to ethical standards.
4. \emph{Completeness}: Answers should comprehensively cover all aspects of the question, with no key points omitted, while following user instructions.
5. \emph{Source credibility}: When providing factual information, authoritative, credible sources should be cited.
6. \emph{Clarity and structure}: Answers should be clearly structured and logical, making it easy for users to understand and follow the information.
7. \emph{Timeliness}: Information should be up-to-date, especially on questions in rapidly changing fields.
8. \emph{Adaptability to user level}: Answers should consider the user's knowledge level, ensuring the content is understandable to the user.

\etitle{(2) Open QA} Open dialogue instructions, usually asking an open-field question, and responses are also open-ended, such as casual chats, advice consultations, recommendations, etc.

Judge criteria.   
1. \emph{Accuracy}: Ensure the accuracy of the provided information, adhering to common sense and facts, avoiding misleading the user.
2. \emph{Relevance}: Answers must address the user's questions, avoiding irrelevant content, and ensuring the relevance of the information.
3. \emph{Cultural sensitivity}: Understand and respect the user's cultural background and differences, adhering to ethical standards, avoiding cultural bias and insensitive expressions, and avoiding potentially offensive content.
4. \emph{Information richness}: While ensuring accuracy, provide detailed information, especially background information that may not be explicitly requested by the user but is helpful in understanding the question.
5. \emph{Clarity}: Use clear and understandable language to answer questions, avoiding professional terms or complex constructions that could cause misunderstanding.
6. \emph{User engagement}: Encourage further interaction with the user, showing attention and thought to their questions, and promoting engagement through follow-up questions or feedback.
7. \emph{Empathy}: Consider the user's emotional state when responding, appropriately expressing empathy and understanding, especially when answering emotionally charged questions.
8. \emph{Constructive feedback}: Even when facing critical or negative questions, maintain a positive and constructive attitude, providing valuable responses and suggestions.

\etitle{(3) Math-related QA} Solve a problem involving mathematics, calculations, reasoning, etc., and the problem has a standard/reference answer.

Judge criteria.   
1. \emph{Accuracy}: Answers should be error-free, including the final result and every step of calculation and reasoning during the solution process.
2. \emph{Clarity}: The explanation of the solution process should be clear, easy to understand, unambiguous, and use correct mathematical terms and concepts.
3. \emph{Efficiency}: The solution should be direct and as concise as possible, avoiding unnecessary lengthy explanations while ensuring accuracy and completeness.
4. \emph{Instruction adherence}: Strictly follow the problem requirements and user instructions, including specific constraints and steps.
5. \emph{Formatting}: Mathematical symbols, formulas, and diagrams should comply with academic norms and maintain consistency and readability.
6. \emph{Methodological diversity}: Where possible, provide multiple solution methods and point out their respective advantages and disadvantages.
7. \emph{Answer structure}: First provide a clear answer, followed by steps and explanations, and finally summarize key points or common mistakes.

\stitle{Writing scenarios}

\etitle{(4) Creative writing} Writing that primarily expresses personalized imagination and emotions, focusing on literary quality and originality, such as creating essays, poems, lyrics, scripts, stories, speeches, social media posts, blogs, advertising materials, brainstorming, etc.

Judge criteria.   
1. \emph{Originality}: The work should reflect the author's independent thinking, include original views and ideas, avoiding plagiarism.
2. \emph{Emotional expression}: The work should effectively convey the author's emotions, resonating with the reader.
3. \emph{Creativity}: The work should exhibit creativity, including unique thinking, language use, and plot construction.
4. \emph{Cultural sensitivity}: The work should respect and consider cultural diversity, avoiding cultural bias, and ensuring the content is free from offensive, inappropriate, or discriminatory material.
5. \emph{Text coherence}: The work should be fluid and logically coherent, with a clear storyline for narrative works.
6. \emph{Stylistic adaptability}: The work should match its style or genre with appropriate writing style and language use.
7. \emph{Imagery and language}: The work should engage readers with expressive language and powerful imagery, enhancing visual and sensory experiences.
8. \emph{User-friendly}: The work should consider the background of the target readers, ensuring content is accessible and readable.

\etitle{(5) Informative and professional writing} Writing aimed at conveying key information and professional knowledge, focusing on accuracy, reliability, and authority, covering practical emails, job applications, product descriptions, user manuals, to in-depth academic papers, medical research, legal opinions, engineering design, industry analysis, economic forecasts, and other complex documents.

Judge criteria.   
1. \emph{Accuracy}: Content must be based on facts and reliable data, reflecting the latest research and real-world conditions to ensure accuracy.
2. \emph{Credibility}: Content should demonstrate authority in its scientific or professional field, supported by reliable sources to build credibility, and provide a detailed assessment framework to increase objectivity.
3. \emph{Harmlessness}: Writing content should meet ethical standards, avoid offensive, inappropriate, or discriminatory material, ensuring the text's harmlessness.
4. \emph{Clarity and coherence}: Information transmission should be clear, accurate, logically rigorous, and sound in arguments, ensuring all readers can easily and orderly understand the conveyed information.
5. \emph{Relevance}: Content should be directly relevant to the topic, focused on the target audience's needs and purposes, avoiding irrelevant information, and ensuring content is valuable to the audience.
6. \emph{Professionalism}: The text should correctly use professional terms and concepts, reflecting the author's deep knowledge and skills in the relevant field, and clearly articulated to suit readers of different levels.
7. \emph{Originality}: The text should reflect the author's original thinking, showcasing unique research, insights, or analysis, ensuring the content's uniqueness.
8. \emph{Formatting standards}: Documents should follow appropriate formatting and design standards, using appropriate professional terms, typesetting, and design elements to enhance content clarity and professional presentation.
9. \emph{Audience engagement}: The text should consider the audience's needs and interests, attract readers through content appeal, effective communication, and interactivity, expanding its impact and response among the audience.

\etitle{(6) Rewriting}  Includes text simplification, language optimization, rewriting text according to instructions, text correction, text summarization and expansion, etc.

Judge criteria.   
1. \emph{Accuracy}: The rewritten text should faithfully convey the original information, avoiding misleading content.
2. \emph{Compliance}: The rewriting should strictly follow the key steps and specific constraints required by the instructions.
3. \emph{Harmlessness}: The text should avoid offensive, inappropriate or discriminatory content.
4. \emph{Text quality}: The text should be grammatically correct, free of spelling errors, and maintain consistency in style, tone, and information.
5. \emph{Relevance}: The rewritten text should be relevant to the target audience and context, ensuring adaptability and targeting.
6. \emph{Conciseness}: The text should be concise and clear, avoiding unnecessary redundancy to convey information clearly.
7. \emph{Originality}: The rewritten text should demonstrate originality, avoiding plagiarism, and providing unique insights or expressions.
8. \emph{Cultural sensitivity}: The text should consider diverse cultural backgrounds, respecting different cultural values and expression habits.

\stitle{Professional scenarios}

\etitle{(7) Translation} Translate the given text into another language without changing the original meaning.

Judge criteria.   
1. \emph{Faithfulness}: The translation needs to maintain fidelity to the original content, ensuring accurate transmission of information, style, and cultural connotations, avoiding misunderstandings.
2. \emph{Fluency}: The translation should be natural and fluent, conforming to the linguistic habits of the target language and easy for readers to understand.
3. \emph{Accuracy}: Terminology and factual information in the translation should be accurate, especially professional terms and data.
4. \emph{Adaptability}: The translation should be adaptively adjusted according to different contexts and target audiences.
5. \emph{Coherence}: The translation should maintain internal logical consistency, ensuring the information's consistency throughout the text.
6. \emph{Cultural appropriateness}: The translation should respect and convey the original text's cultural characteristics while considering the target language's cultural acceptance.
7. \emph{Harmlessness}: The translation should avoid any potentially misleading or offensive content, ensuring cultural and contextual sensitivity.
8. \emph{Innovation}: Without violating the original meaning, the translation should demonstrate creativity, making the translation dynamic and closer to the target language's expression habits.

\etitle{(8) Reading comprehension and extraction} Read materials and complete directive tasks based on the materials, such as Q\&A, summarization, keyword extraction, topic extraction, title generation, fact-checking, etc.

Judge criteria.   
1. \emph{Accuracy}: Answers should strictly correspond to the given context information, correctly responding to the questions, even if the context information might be incorrect or outdated.
2. \emph{Relevance}: Answers should directly correspond to the text content or topic, avoiding irrelevant information, ensuring all provided information has a clear textual or thematic basis.
3. \emph{Instruction compliance}: The output should strictly follow the specific requirements of the instructions, including action steps and any constraints.
4. \emph{Text coherence}: Answers should have internal logical consistency and fluency, ensuring the consistency and coherence of the information.
5. \emph{User experience}: Answers should be presented in a user-friendly manner, easy to understand, and guide the user to obtain the needed information timely.
6. \emph{Contextual understanding}: The model should demonstrate the ability to understand complex contexts and implicit information.
7. \emph{Conciseness}: Information expression should be direct and concise, avoiding unnecessary elaboration or complexity.
8. \emph{Creativity}: In tasks requiring creative output (such as title or summary generation), answers should exhibit a certain degree of originality and appeal.

\etitle{(9) Role-playing} Pretend to be a particular person, character, profession, or identity, and complete the tasks in the instructions based on this role.

Judge criteria.   
1. \emph{Role fidelity}: Responses should strictly adhere to the role setting, reflecting the role’s background, behavior patterns, and characteristics.
2. \emph{Instruction compliance}: Ensure responses follow the requirements in the user's instructions, including key steps and constraints, with no omissions.
3. \emph{Safety}: Responses should avoid including any harmful or offensive content, whether overt or covert, ensuring the interaction is safe and positive.
4. \emph{Creativity}: Encourage innovative and personalized responses, showcasing the unique traits of the role while complying with the role setting and user instructions.
5. \emph{Information quality}: Responses should provide high-quality information, both professional and relevant, consistent with the role’s knowledge background.
6. \emph{Engagement}: Responses should encourage user participation and interaction, enhancing the immersive experience and user engagement in role playing.
7. \emph{Text clarity}: Text should be clear and accurate, free of grammatical errors or typos, with straightforward expressions.
8. \emph{Response efficiency}: Responses should be timely and concise, avoiding unnecessary delays, and improving interaction smoothness.

\etitle{(10) Programming-related}  Tasks related to computer code, including implementing code based on requirements, code modification and optimization, programming language conversion, analyzing code and responding to related questions, software development assistance, education, and learning, etc.

Judge criteria.   
1. \emph{Code correctness}: Code should strictly follow the requirements or specifications, free from syntax or logic errors, achieving the intended functionality.
2. \emph{Code maintainability}: Code should be easy to understand and modify, with good modularity, documentation comments, and adherence to coding standards.
3. \emph{Security and safety}: Code should not contain any potential security vulnerabilities or malicious behaviors, handle all types of input safely, and have a stable error-handling mechanism.
4. \emph{Performance efficiency}: Code should execute efficiently, with optimized resource utilization, considering the algorithm's complexity and data structure choice.
5. \emph{Analysis accuracy}: Code analysis should be thorough and accurate, correctly understanding the code logic, data flow, and structure.
6. \emph{Modularity}: The written code should be modular, clearly separating concerns. It should use appropriate functions, classes, and modules to facilitate reusability and maintainability.
7. \emph{Comprehensibility of explanations}: Explanations and analyses of the code should be clear, easy to understand by users, with appropriate terms and language style.
8. \emph{Problem-solving effectiveness}: In tasks involving code modification, optimization, and programming language conversion, the generated or analyzed code should effectively solve the specified problem.
9. \emph{Input/output adherence}: Code should handle input and output in accordance with user-specified requirements, including format, type, and size. In the absence of specific instructions, it should follow common standards.

\begin{table}[tbh!]
  \centering
  \caption{Scenario mapping between \modelname and Llama-3}
  \label{tab:scenarioComp}
  \begin{tabularx}{.47\textwidth}{X X}
    \toprule
    \textbf{\modelname scenarios} & \textbf{Llama-3 use cases} \\ \midrule
    (1*) Closed-QA & (1) Closed question answering \\ \midrule
    \multirow{3}{*}{(2*) Open-QA} & (2) Open question answering  \\
        & (3) Asking for advice  \\ 
        & (4) Brainstorming \\ \midrule
    (3) Math-related QA & (5) Reasoning  \\ \midrule 
    (4*) Creative writing & (6) Creative writing  \\ \midrule
    (5) Info. and prof. writing &  \texttt{NA} \\ \midrule
    (6*) Rewriting & (7) Rewriting  \\ \midrule
    (7) Translation & \texttt{NA} \\ \midrule
    \multirow{2}{*}{(8*) Reading compre. and extrac.} & (8) Extraction  \\
    & (9) Summarization  \\ \midrule
    (9*) Role-playing & (10) Inhabiting a character  \\ \midrule
    (10*) Programming-related & (11) Coding  \\ \midrule
    \texttt{NA} & (12) Classification  \\ 
    \bottomrule
  \end{tabularx}
\end{table}

\stitle{Justification for scenario design}
Table~\ref{tab:scenarioComp} illustrates the mapping between \modelname's scenarios and Llama-3's evaluation use cases~\cite{llama3tech}. 
From the table we can see that 7 out of \modelname's 10 scenarios, \ie those with *, have direct mapping to the ones of Llama-3. The remaining math-related QA, informative and professional writing, and translation are popular applications of LLMs and deserve context-specific evaluations.
On the other side, 11 out of Llama-3's 12 use cases, except for classification, are covered by \modelname.
As the scenario design of \modelname is independent with Llama-3, the above comparison could be regarded as an empirical justification for our scenario design through human-AI collaboration.


\begin{table*}[tbh!]
  \caption{Prompt template for reference-guided grading. Difference with Table~\ref{tab:prompt} have been emphasized in bold.}
  \label{tab:prompt-with-ref}
  \small
  \begin{tabularx}{\textwidth}{X}
    \toprule
        \emph{The same task description as in Table~\ref{tab:prompt}.}

        \ 
        
        \emph{The same grading guidelines as in Table~\ref{tab:prompt}.}
        
        \
        
        Regarding a user instruction of [\texttt{\{scenario name\}}] \textbf{and the corresponding reference answer}, we have collected the following AI assistant response. \textbf{Based on your understanding of the current evaluation standards, compare the reference answers and evaluate these responses comprehensively.  Note that the reference answer may not be the only possible one; it is merely used to demonstrate the standard for a high-level (specifically, at the 4th tier) response.} Below are the user instruction, \textbf{reference answer}, and the assistant's response data: 
        
        [Data Begin] 
        
        ***
        
        [User Instruction]: \texttt{\{instruction\}} 
        
        *** 

        \textbf{[Reference answer]: \texttt{\{reference answer\}}} 
        
        ***
        
        [Response]: \texttt{\{response\}}
        
        ***
        
        [Data End] 

        \
        
        You need to follow these steps to evaluate the above response: 
        
        1. Recall the relevant AI assistant response criteria and carefully read and understand the response to be evaluated.
        
        2. Identify from all criteria the key ones for the current user instruction and response, including those that performed well and those that did not. 
        
        3. Besides the given criteria, add any other important criteria that you think are necessary for evaluating the current user instruction response. 
        
        4. Based on your final selection of criteria, \textbf{compare the reference answer and} assign scores (between 1-5) to each criterion, and provide a comprehensive score after weighting all sub-scores. 

        \ 
        
        \emph{The same output requirement and format as in Table~\ref{tab:prompt}.}  \\
    \bottomrule
  \end{tabularx}
\end{table*}

\begin{table*}[tbh!]
  \caption{Prompt template for pairwise comparison. Difference with Table~\ref{tab:prompt} have been emphasized in bold.}
  \label{tab:prompt-pairwise}
  \small
  \begin{tabularx}{\textwidth}{X}
    \toprule
        \emph{The same task description as in Table~\ref{tab:prompt}.}

        \ 
        
        \emph{The same grading guidelines as in Table~\ref{tab:prompt}.}
        
        \
        
        Regarding a user instruction of [\texttt{\{scenario name\}}] , we have collected \textbf{two responses from AI assistants. Based on your understanding of the current standards for responses, comprehensively evaluate and determine which response is better or if they are tied (including both being good or both being bad).} Below are the user instruction and the assistant's response data: 
        
        [Data Begin] 
        
        ***
        
        [User Instruction]: \texttt{\{instruction\}} 
        
        *** 
        
        \textbf{[Response 1]: \texttt{\{response 1\}}}
        
        ***

        \textbf{[Response 2]: \texttt{\{response 2\}}}

        ***
        
        [Data End] 

        \
        
        \emph{The same evaluation steps as in Table~\ref{tab:prompt}.}

        \ 
        
        Think carefully and then provide your conclusion. Your response should keep the `[[' and `]]' symbols in the output: 
        
        I believe \textbf{[[Response 1 is better]]/[[Response 2 is better]]/[[Both Responses are tied]], with the overall score for Response 1 being [[a score between 1 and 5]], and the overall score for Response 2 being [[a score between 1 and 5]], based on the following reasons:}
        
        \textbf{1. (Please detail your reasons in order of importance from high to low, each standard also attaching the [[scores]] for both responses under that standard...)}   \\
    \bottomrule
  \end{tabularx}
\end{table*}

\section{Prompts}
\label{app:prompts}

Prompts related to \modelname training and inference are summarized in this section, including the ones for reference-guided grading (Table~\ref{tab:prompt-with-ref}), pairwise comparison (Table~\ref{tab:prompt-pairwise}), reference-based question synthesis (Table~\ref{tab:prompt-qllm-train}), role-playing quizzing for three scenarios (Table~\ref{tab:prompt-rpq-math}--\ref{tab:prompt-rpq-reading}), and fint-tuning the sceanrio classification LLM (Table~\ref{tab:prompt-scenario-classification}). Note that \{\texttt{variable}\} represents a variable and should be filled in properly for all prompts. 

We also illustrate a complete evaluation record for supervised fine-tuning, with the evaluation input in Table~\ref{tab: math example} and the evaluation result by GPT-4 in Table~\ref{tab: gpt-4's response}. Evaluations by Qwen-max, Qwen-14B, \modelname are also presented (Tables~\ref{tab: qwenmax's response}--\ref{tab: themis' response}). Among the three, we find only \modelname makes a reasonable evaluation.

\begin{table*}[tbh!]
  \caption{Prompt template for question synthesis to train the questioning model.} 
  \label{tab:prompt-qllm-train}
  \small
  \begin{tabularx}{\textwidth}{X}
    \toprule
        Requirements for the scenario:

        Name: \{\texttt{scenario name}\}
        
        Definition: \{\texttt{scenario description}\}
        
        \ 
        
        Reference Text:
        
        \{\texttt{reference text}\}

        \ 
        
        Requirements:
        
        1. The generated questions and answers should be based on the article content and should meet the scenario requirements.
        
        2. Questions should be detailed, containing necessary information to encourage thorough answers.
        
        3. If the information in the reference text is insufficient to generate question-answer pairs, return the following: "Sorry, this article does not contain enough information related to \{\texttt{scenario name}\} to generate relevant questions and answers."

        4. The generated question-answer pairs need to simulate questions and answers people might consult the LLM about in real-life scenarios.
        
        5. Ensure the completeness and answerability of the questions independently; include the original content if necessary.
        
        6. Ensure the correctness of the answers.
        
        \ 
        
        Sample Questions:
            
            Example 1: \{\texttt{example 1}\}
            
            Example 2: \{\texttt{example 2}\}
            
            Example 3: \{\texttt{example 3}\}

        \

        Please generate 5 sets of question-answer pairs that meet the requirements:
        
        QUESTION: [The generated question based on article content]
        
        ANSWER: [The answer to the question, if possible, based on article content; otherwise, based on model's own knowledge]
        
        LEVEL: [The difficulty level of the question: easy/medium/difficult]
        
        [END OF QA PAIR] \\
    \bottomrule
  \end{tabularx}
\end{table*}

\begin{table*}[tbh!]
  \caption{Prompt template for question generation with role-playing quizzing: math-related QA.}
  \label{tab:prompt-rpq-math}
  \small
  \begin{tabularx}{\textwidth}{X}
    \toprule
        I need to create exam questions to assess students' proficiency in mathematics. Please help me generate 10 \{\texttt{difficult level}\} questions for \{\texttt{audience}\} in the subject of \{\texttt{subject}\}. Each question should contain varying numbers of unknowns or mathematical symbols. Ensure the language, clarity, and accuracy of the questions: the problems must be described in \{\texttt{language}\} and be clear and unambiguous with precise definitions to ensure solvability and that a standard answer exists. Please strictly follow my instructions regarding the number of questions.

        \ 
        
        Please output all generated questions in jsonl format, with one question per line in a JSON description, and do not output any additional content. Here is an example of a question:
        
        \{"question": "[the generated question]", "level": "[the difficult level of the question]", "subject": "[the subject of the question]"\}  \\
    \bottomrule
  \end{tabularx}
\end{table*}

\begin{table*}[tbh!]
  \caption{Prompt template for question generation with role-playing quizzing: programming-related.}
  \label{tab:prompt-rpq-code}
  \small
  \begin{tabularx}{\textwidth}{X}
    \toprule
        You are a senior recruitment expert with vast hands-on programming experience at 
        [Company: \{\texttt{company}\}]. For the upcoming recruitment season, you are designing a series of unique questions to assess candidates' programming skills. Now, please design 10 \{\texttt{difficult level}\} programming or code analysis questions for \{\texttt{audience}\}, under the theme of \{\texttt{topic}\}. Ensure each question is described in \{\texttt{language}\}, with clear, unambiguous language and well-defined concepts, and that content (code, description, etc.) is complete. Ensure the questions are solvable with a standard answer. The questions or answers should directly involve programming code, with the code ranging from 30 to 50 lines, neither too simple nor too difficult, to be solvable within a reasonable time. When necessary, provide clear code snippets in the questions.

        \ 
        
        Please output all generated questions in jsonl format, with one question per line in a JSON description, and do not output any additional content. Here is an example of a question:

        \{"question": "[the generated question]", "company": "[the company]", "level": "[the question difficult level]", "topic": "[the question topic]"\}

        Stick strictly to your role, think for a moment, and then start drafting the questions. \\
    \bottomrule
  \end{tabularx}
\end{table*}

\begin{table*}[tbh!]
  \caption{Prompt template for question generation with role-playing quizzing: reading comprehension and extraction.}
  \label{tab:prompt-rpq-reading}
  \small
  \begin{tabularx}{\textwidth}{X}
    \toprule
        You are now a language expert preparing reading comprehension and extraction questions for a language proficiency test. The task definition is: read materials and complete directive tasks based on the materials, such as Q\&A, summarization, keyword extraction, topic extraction, title generation, fact-checking, etc.
        Please prepare 10 questions based on the following reading materials. The questions need to cover various task types, such as: [title generation, summary, theme keyword extraction], [key information and concept Q\&A], [content understanding, explanation, and fact-checking], [content comparison, analysis, and summary], [critical thinking and secondary creation]. Ensure a relatively balanced distribution across different task types. Ensure each question is described in \{\texttt{language}\}, clear and unambiguous with well-defined concepts, and the description is complete. Ensure the questions are solvable without exceeding the scope of the reading materials.
        
        \ 
        
        Please output all generated questions in jsonl format, with one question per line in a JSON description, in the following format:
        
        \{"question": "[the generated question]", "answer": "[your answer to the question]", "task": "[the task type]"\}

        Stick strictly to your role, think for a moment, and then start drafting the questions. The reading material is as follows:
        
        \{\texttt{reading material}\}\\
    \bottomrule
  \end{tabularx}
\end{table*}

\begin{table*}[tbh!]
  \caption{Prompt template for fine-tuning the scenario classification LLM.}
  \label{tab:prompt-scenario-classification}
  \small
  \begin{tabularx}{\textwidth}{X}
    \toprule
        User queries for large language models can generally be categorized into the following 10 scenarios:

        1. Close QA: Solve a problem that may involve professional knowledge or real-world inquiries, such as historical facts or scientific laws, and the problem has a standard/reference answer.

        2. Open QA: Open dialogue instructions, usually asking an open-field question, and responses are also open-ended, such as casual chats, advice consultations, recommendations, etc.

        3. Math-related QA: Solve a problem involving mathematics, calculations, reasoning, etc., and the problem has a standard/reference answer.

        4. Creative writing: Writing that primarily expresses personalized imagination and emotions, focusing on literary quality and originality, such as creating essays, poems, lyrics, scripts, stories, speeches, social media posts, blogs, advertising materials, brainstorming, etc.
        
        5. Informative and professional writing: Writing aimed at conveying key information and professional knowledge, focusing on accuracy, reliability, and authority, covering practical emails, job applications, product descriptions, user manuals, to in-depth academic papers, medical research, legal opinions, engineering design, industry analysis, economic forecasts, and other complex documents.
        
        6. Rewriting:  Includes text simplification, language optimization, rewriting text according to instructions, text correction, text summarization and expansion, etc.
        
        7. Translation: Translate the given text into another language without changing the original meaning.
        
        8. Reading comprehension and extraction: Read materials and complete directive tasks based on the materials, such as Q\&A, summarization, keyword extraction, topic extraction, title generation, fact-checking, etc.
        
        9. Role-playing: Pretend to be a particular person, character, profession, or identity, and complete the tasks in the instructions based on this role.
        
        10. Programming-related:  Tasks related to computer code, including implementing code based on requirements, code modification and optimization, programming language conversion, analyzing code and responding to related questions, software development assistance, education, and learning, etc.
        
        \ 
        
        Now I have a user query as follows:
    
        [\{\texttt{user instruction}\}]
 
        Please determine which scenario this query belongs to based on the aforementioned 10 scenarios (if you cannot determine, you can classify it as "default").

        Please directly provide the name of the scenario.\\ \midrule
        \{\texttt{labeled scenario}\} \emph{Note: This output will be used for fine-tune.} \\ 
    \bottomrule
  \end{tabularx}
\end{table*}

\begin{table*}[tbh!]
  \caption{A complete prompt about answering mathematical questions.}
  \label{tab: math example}
  \small
  \begin{tabularx}{\textwidth}{X}
    \toprule
    Input \\ \midrule
    Your task is to rate the quality of responses provided by the AI intelligent assistant.\\
    You are very clear that when a user issues an instruction in a scenario about "answering mathematical questions" (defined as: solving a problem involving mathematics, calculation, reasoning, etc., with a standard/reference answer), the response of an AI intelligent assistant should meet the following standards (listed from most to least important):\\
    
    [Standards Start]\\
    
    1. Accuracy: The response should be accurate, including every step of calculations and reasoning involved in the solving process.\\
    2. Clarity: The explanation of the solving process should be clear, easy to understand, unambiguous, and use mathematical terms and concepts correctly.\\
    3. Efficiency: The response should be direct and as concise as possible, avoiding unnecessary lengthy explanations, while ensuring accuracy and completeness.\\
    4. Command Compliance: Strictly follow the problem requirements and user instructions, including handling specific restrictions and steps.\\
    5. Formatting Standards: The use of mathematical symbols, formulas, and diagrams should adhere to academic norms and be consistent and readable.\\
    6. Method Diversity: Where possible, provide multiple solving methods and indicate their respective pros and cons.\\
    7. Answer Structure: Present a clear answer first, followed by steps and explanations, and finally summarize key points or common mistakes.\\
    
    [Standards End]\\
    
    The scoring is on a 5-level scale (1-5), with each score level meaning the following:\\
    
    [Score Levels Start]\\
    
    1 Response has major flaws, completely diverts from standards, and should not appear in practice;\\
    2 Response has some content that meets the standards, can be adopted, but as a whole, the quality of the response is not acceptable;\\
    3 Response has both pros and cons, overall pros outweigh the cons within the required evaluation standards;\\
    4 Response quality is satisfactory, generally meets the standards, with a few minor issues that could be improved. This level represents the response quality as shown in the given reference answer;\\
    5 Response is perfect in all aspects, strictly adheres to the standards. This level represents a response quality better than the reference answer.\\
    
    [Score Levels End]\\
    
    \\
    In response to a user instruction in an "answering mathematical questions" scenario, we collected the following reply from an AI intelligent assistant.\\
    Please assess the response based on your knowledge of the current scenario’s standards for intelligent assistant responses, and provide an evaluation. Here are the user instruction and assistant response data:\\
    
    [Data Start]\\
    
    ***\\
    
    [User Instruction]: Which is greater, 9.11 or 9.9 \\
    
    ***\\
    
    [Response]: First, we need to compare the two numbers 9.9 and 9.11.\\
    
    \\
    The integer part of 9.9 is 9.\\
    The integer part of 9.11 is also 9.\\
    Since the integer parts are the same, we need to compare the decimal parts.\\
    \\
    The decimal part of 9.9 is 0.9.\\
    The decimal part of 9.11 is 0.11.\\
    Comparing the decimal parts, 0.9 is greater than 0.11.\\
    \\
    \textcolor{red}{Therefore, 9.11 is greater than 9.9.}\\
    ***\\
    
    [Data End]\\
    
    \\
    You need to follow the process below to assess the above response:\\
    1. Recall the relevant standards for intelligent assistant responses, and carefully read and understand the response to be assessed.\\
    2. Identify key standards from all standards for the current user instruction and response, including good and insufficient points.\\
    3. In addition to the given standards, add other important standards you think are necessary for evaluating the current user instruction response.\\
    4. Based on your final selection of standards, score the response (between 1-5), and after weighting all minor scores, give an overall score for the response.\\
    Think carefully for a moment, then give your conclusion. Your return template should be as follows, note to retain the '[' and ']' in the output:\\
    ***\\
    I believe the overall score of this response is [[a score between 1-5]], for the following reasons.\\
    Current response strengths:\\
    1. (Please list in sequence what you think are the strengths of the current response, and give [[a score between 1-5]] for each...)\\
    Current response weaknesses:\\
    1. (Please list in sequence what you think are the weaknesses of the current response, and give [[a score between 1-5]] for each...)\\
    ***\\ 
    \bottomrule
  \end{tabularx}
\end{table*}

\begin{table*}[tbh!]
  \caption{Evaluation result from GPT-4 to the prompt in Table~\ref{tab: math example}.}
  \label{tab: gpt-4's response}
  \small
  \begin{tabularx}{\textwidth}{X}
    \toprule
        GPT-4's response \\ \midrule

        I believe the overall rating for this reply is [[1]] for the following reasons:\\
        Advantages of the current reply:\\
        1. Clarity: The explanation process is relatively clear, comparing integer parts and decimal parts in steps, which is easy to understand. [[3]]\\
        \\
        Shortcomings of the current reply:\\
        1. Accuracy: The final conclusion of the reply is incorrect; in fact, 9.9 is greater than 9.11, not as stated in the reply that "9.11 is greater than 9.9." [[1]]\\
        2. Efficiency: Although a comparison process is provided, the efficiency becomes meaningless after presenting an incorrect conclusion. [[1]]\\
        3. Instruction Compliance: It fails to correctly follow the command to provide an accurate answer and does not provide the correct comparison result as per the user’s request. [[1]]\\
        4. Method Diversity: It does not offer different comparison methods, such as direct numerical comparison, and only uses a part-by-part comparison approach. [[2]]\\
        5. Answer Structure: Although the structure is clear, comparing integer parts first and then decimal parts, this structure has not effectively served to present the correct answer due to the incorrect final answer. [[1]]\\
        \\
        Considering the above ratings, due to the core issue of accuracy, even though other aspects are passable, the overall rating remains at the lowest tier [[1]].\\
    \bottomrule
  \end{tabularx}
\end{table*}

\begin{table*}[tbh!]
  \caption{Evaluation result from Qwen-Max to the prompt in Table~\ref{tab: math example}.}
  \label{tab: qwenmax's response}
  \small
  \begin{tabularx}{\textwidth}{X}
    \toprule
        Qwen-Max's response \\ \midrule

        I believe that the overall rating for this reply is [[3]], for the following reasons.\\
        \\
        Strengths of the current reply:\\
        1. Accuracy: Although the final conclusion is incorrect, the decomposition of numbers during the problem-solving process is quite clear, showing a comparative thought process. [[3]]\\
        2. Clarity: The steps of the solution are clearly articulated, and the logical sequence is easy to understand. [[4]]\\
        3. Compliance with instructions: It directly answered the user's instruction to compare the sizes of two numbers. [[4]]\\
        \\
        Weaknesses of the current reply:\\
        1. Accuracy: The final conclusion is erroneous, stating that 9.11 is larger than 9.9 is incorrect; \textcolor{red}{in reality, 9.9 is less than 9.11}. This is a core error and significantly affects the rating. [[1]]\\
        2. Formatting standards: Although no complex mathematical symbols or formulas were used, the basic comparison symbols should be used correctly. [[2]]\\
        3. Answer structure: The structure of the answer, steps, and explanation is fundamentally reasonable, but the incorrect conclusion impacts the overall effectiveness of the structure. [[2]]\\
        \\
        Taking everything into consideration, despite the reply’s acceptable performance in terms of clarity and following instructions, the core error in accuracy cannot be overlooked, therefore a rating of 3 is given for the overall evaluation.\\
    \bottomrule
  \end{tabularx}
\end{table*}

\begin{table*}[tbh!]
  \caption{Evaluation result from Qwen-14B to the prompt in Table~\ref{tab: math example}.}
  \label{tab: qwenturbo's response}
  \small
  \begin{tabularx}{\textwidth}{X}
    \toprule
        Qwen-Turbo's response \\ \midrule
        I believe the \textcolor{red}{overall rating for this response is [[4]]}, and here are the reasons.\\
        \\
        Advantages of the current response:\\
        1. \textcolor{red}{The reply is accurate and precise}, including the final result and each step of calculation and reasoning during the problem-solving process, [[5]].\\
        2. The explanation of the problem-solving process is clear, easy to understand, unambiguous, and uses mathematical terms and concepts correctly, [[5]].\\
        3. The answer is direct and concise, avoiding unnecessary lengthy explanations, while ensuring accuracy and completeness, [[4]].\\
        4. Strictly followed the requirements of the question and user instructions, including dealing with specific constraints and steps, [[5]].\\
        5. The use of mathematical symbols, formulas, and diagrams adheres to academic standards and maintains consistency and readability, [[5]].\\
        \\
        Shortcomings of the current response:\\
        1. Fails to provide multiple methods of solving the problem where possible, nor does it discuss the pros and cons of each, [[2]].\\
        2. The response lacks a structured answer format, that is, presenting a clear answer first, followed by steps and explanations, and concluding with key points or common mistakes, [[2]].\\
        \\
        In summary, this response excels in accuracy, clarity, efficiency, and adherence to instructions, but lacks in methodological diversity and structured answer format. Therefore, I have given it an overall rating of [[4]].\\
    \bottomrule
  \end{tabularx}
\end{table*}

\begin{table*}[tbh!]
  \caption{Evaluation result from Themis to the prompt in Table~\ref{tab: math example}.}
  \label{tab: themis' response}
  \small
  \begin{tabularx}{\textwidth}{X}
    \toprule
    Themis' response \\ \midrule
    I believe the overall rating for this reply is [[2]] for the following reasons:\\
    Advantages of the current reply:\\
    1. Clarity: The response is clear in expression, easy to understand, and does not use complex mathematical terms, which helps users understand the comparison process. [[4]]\\
    2. Formatting Norms: Although this response does not directly use mathematical symbols or formulas, the way it explains number comparisons is standard. [[4]]\\
    \\
    Shortcomings of the current reply:\\
    1. Accuracy: The conclusion of the response is incorrect. In fact, 9.9 is greater than 9.11, because although the integer part is the same, the decimal part of 9.9 (0.9) is greater than that of 9.11 (0.11). This mistake directly violates the highest standard of accuracy. [[1]]\\
    2. Efficiency: Although the response attempts to directly compare two numbers, this attempt does not achieve the efficiency it should due to the incorrect conclusion. [[2]]\\
    3. Compliance with Instructions: The response does not correctly follow the user's instructions, as the user asked which number is larger, and the response provided an incorrect answer. [[1]]\\
    4. Method Diversity: The response does not provide a variety of methods to solve the problem, though it may not be necessary for this simple issue. However, considering the mistake in the response, this also becomes a drawback. [[2]]\\
    5. Answer Structure: Although the response attempts to first give an answer and then explain it, because the answer is incorrect, this structure does not achieve its intended effect. [[1]]\\
    \\
    In conclusion, although this response performs well in Clarity and Formatting Norms, it falls short in Accuracy, Efficiency, Compliance with Instructions, Method Diversity and Answer Structure. Therefore, it receives an overall rating of 2.\\
    \bottomrule
  \end{tabularx}
\end{table*}
